\documentclass[Afour,sageh,times]{sagej}
\usepackage{moreverb,url}
\usepackage{tabularx}
\usepackage{booktabs}
\usepackage[utf8]{inputenc}
\usepackage[utf8]{inputenc}  % doit être AVANT
\usepackage{csquotes}
\usepackage[colorlinks,bookmarksopen,bookmarksnumbered,citecolor=red,urlcolor=red]{hyperref}
\usepackage{subcaption}
\newcommand\BibTeX{{\rmfamily B\kern-.05em \textsc{i\kern-.025em b}\kern-.08em
T\kern-.1667em\lower.7ex\hbox{E}\kern-.125emX}}

\def\volumeyear{2016}

\begin{document}

\runninghead{Kabal et al.}

\title{A Unified Benchmark for Evaluating Knowledge Graph Construction Methods and Graph Neural Networks}

\author{Othmane Kabal\affilnum{1}, 
Mounira Harzallah\affilnum{1}, 
Fabrice Guillet\affilnum{1}, 
Hideaki Takeda\affilnum{2}, 
Ryutaro Ichise\affilnum{3}}

\affiliation{
\affilnum{1}Nantes University, LS2N, Nantes, 44300, France\\
\affilnum{2}National Institute of Informatics, Chiyoda-ku, Tokyo, 101-8430, Japan\\
\affilnum{3}Institute of Science Tokyo, Tokyo, 152-8550, Japan
}

\corrauth{Othmane Kabal, Nantes University, LS2N, 
44300 Nantes, France.}

\email{othmane.kabal@univ-nantes.fr}

\begin{abstract}
Knowledge graphs automatically constructed from text are increasingly used in real-world applications. However, their inherent noise, fragmentation, and semantic inconsistencies significantly affect the performance of Graph Neural Networks (GNNs) on downstream tasks. Assessing their performance and robustness remains difficult, as it is often unclear whether observed results stem from the learning model or from the quality of the constructed graph itself.
In this work, we introduce a dual-purpose benchmark designed to jointly evaluate (i) the performance of GNNs on noisy, text-derived graphs and (ii) the effectiveness of graph construction methods on a downstream task. The benchmark is built in the biomedical domain from a single textual corpus and includes two automatically constructed graphs generated using different extraction methods, alongside a high-quality reference graph curated by experts that serves as an upper performance bound. This design enables controlled comparison of construction methods and systematic evaluation of GNN robustness through semi-supervised node classification.
We further provide a standardized, reproducible, and extensible evaluation framework, facilitating the integration of new graph extraction methods and learning models.
% Code and resources are publicly available at:~\url{https://github.com/OthmaneKabal/text_driven_kg_bench}
\end{abstract}
\keywords{Knowledge Graphs, Graph Neural Networks, Benchmarking, Text-Derived Knowledge Graphs, Graph Representation Learning, Knowledge Graph Evaluation}
\maketitle

\section{Introduction}
Knowledge graphs (KGs) play an increasingly central role in a wide range of artificial intelligence applications~\cite{ji2021survey, peng2023knowledge}. When combined with Graph Neural Networks (GNNs), they have demonstrated strong performance across numerous tasks, including node classification \cite{xiao2022graph}, relation prediction \cite{arrar2024comprehensive}, and knowledge graph completion \cite{liang2024survey}. The rapid advancement of these models~\cite{10670406}, coupled with the growing availability of textual data and new KG construction methods~\cite{zhong2023comprehensive}, has created a pressing need for benchmarks that can keep pace.\\
However, existing benchmarks exhibit several important limitations. Early and widely used academic datasets such as Citeseer and Cora~\cite{sen2008collective} are relatively small and consist of monorelational graphs, making them poorly suited for evaluating modern GNN architectures designed to operate on large-scale, heterogeneous, and multi-relational knowledge graphs. Even multirelational benchmarks such as AIFB~\cite{ristoski2016collection} or domain-specific datasets in fields such as chemistry~\cite{NEURIPS2023_7f755e27} and biology~\cite{zitnik2017predicting} represent entities primarily as symbolic identifiers or low-dimensional feature vectors, with little or no textual information. As a result, these benchmarks fail to capture the rich semantic content that  is critical for many real-world applications. In contrast, large-scale knowledge graphs such as Wikidata~\cite{10.1145/2629489}, YAGO~\cite{suchanek2024yago}, DBpedia~\cite{auer2007dbpedia}, or domain-specific resources such as the Unified Medical Language System (UMLS)~\cite{bodenreider2004unified} contain extensive textual information associated with entities and relations. These resources are generally considered high-quality, as they are constructed from structured sources or curated by domain experts. While they provide valuable semantic information, they do not adequately reflect the challenges faced by graph learning models when operating on knowledge graphs that are automatically constructed from unstructured text.

Recent advances in natural language processing have enabled the automatic construction of knowledge graphs directly from large collections of unstructured documents~\cite{kabal2024enhancing, anuyah2025automated, mo2025kggen}. These text-derived knowledge graphs offer a more realistic representation of real-world data, as their nodes correspond to natural language terms extracted from text. However, such graphs are inherently noisy~\cite{mihindukulasooriya2017towards, cai2025understanding}. They often exhibit ambiguity, redundancy, fragmentation, and semantic inconsistencies introduced during the extraction process. Moreover, the resulting noise patterns are complex and domain-dependent, and therefore differ substantially from the simplified synthetic noise typically considered in controlled experimental studies~\cite{cai2025understanding}. Despite the growing importance of automatically constructed knowledge graphs, their evaluation remains challenging. Most existing studies focus primarily on the correctness of extracted triples, typically through manual annotation or automatic evaluation metrics~\cite{yu2023compleqa, kabal2024enhancing,lairgi2024itext2kg, mihindukulasooriya2023text2kgbench}. However, such approaches generally assess only a subset of extracted relations and do not capture the overall quality of the resulting graph. Other works adopt a more practical perspective by evaluating the whole constructed graph through downstream tasks~\cite{mo2025kggen, heist2023kgreat, yang2025graphusion}. However, these approaches still suffer from an important limitation: the expected level of performance remains unclear. In particular, the absence of datasets that simultaneously provide (i) a textual corpus, (ii) automatically constructed knowledge graphs, and (iii) a high-quality reference graph makes it difficult to systematically analyze how extraction errors propagate and affect downstream learning performance~\cite{cai2025understanding}. As a result, it becomes challenging to determine whether performance degradation originates from the quality of the constructed graph itself or from the learning method applied on top of it.

To bridge this gap, we introduce a novel dual-purpose benchmark designed to enable the controlled evaluation of both knowledge graph construction methods and GNN models. The benchmark is built on the MedMentions~\cite{mohan2019medmentions} biomedical corpus and includes two knowledge graphs automatically constructed from the same textual source using different extraction pipelines, together with a high-quality reference graph derived from the UMLS thesaurus. Because these graphs share a common set of entities and annotations, they allow direct comparison across graphs of varying quality under identical experimental conditions.
The benchmark supports two complementary objectives. First, it enables the evaluation of knowledge graph construction methods by measuring how the quality of automatically extracted graphs affects performance on a downstream learning task (semi-supervised classification), compared to the performance obtained on the clean reference graph. This allows the performance loss caused by graph construction errors to be explicitly quantified. Second, it provides a standardized framework for evaluating the robustness of GNN models operating on noisy, text-derived knowledge graphs by comparing model performance across the automatically constructed graphs and the clean reference graph, which acts as an upper bound on the performance that can be achieved.
To ensure reproducibility, the benchmark also includes a complete evaluation pipeline, providing standardized data loaders, training protocols, and baseline implementations of several representative GNN architectures.
To summarize, the main contributions of this work are threefold:
\begin{itemize}
    \item A dual-purpose benchmark designed to jointly evaluate (i) knowledge graph construction methods through their impact on a downstream task and (ii) Graph Neural Network models operating on text-derived knowledge graphs under controlled experimental conditions. 
    \item The benchmark itself, which comprises a textual corpus, two automatically constructed KGs from different pipelines, and a high-quality reference graph derived from the UMLS Thesaurus. Importantly, all graphs share a common set of nodes annotated according to the same schema, enabling controlled comparisons across graphs of varying quality.
    \item A reproducible and extensible evaluation framework, providing standardized data loaders, training protocols, and baseline GNN implementations, allowing fair and systematic benchmarking of both graph construction approaches and graph learning models.
\end{itemize}
The remainder of this paper is structured as follows. Section 2 describes the benchmark construction, including the reference graph, the textual corpus, and the automatically generated knowledge graphs. Section 3 presents the evaluation protocol and benchmarking scenarios. Section 4 reports the experimental results obtained with baseline GNN models. Finally, Section 5 concludes the paper and outlines directions for future work.

\section{Benchmark Construction}
\begin{figure*}[ht]
  \centering
  \includegraphics[width=0.8\linewidth]{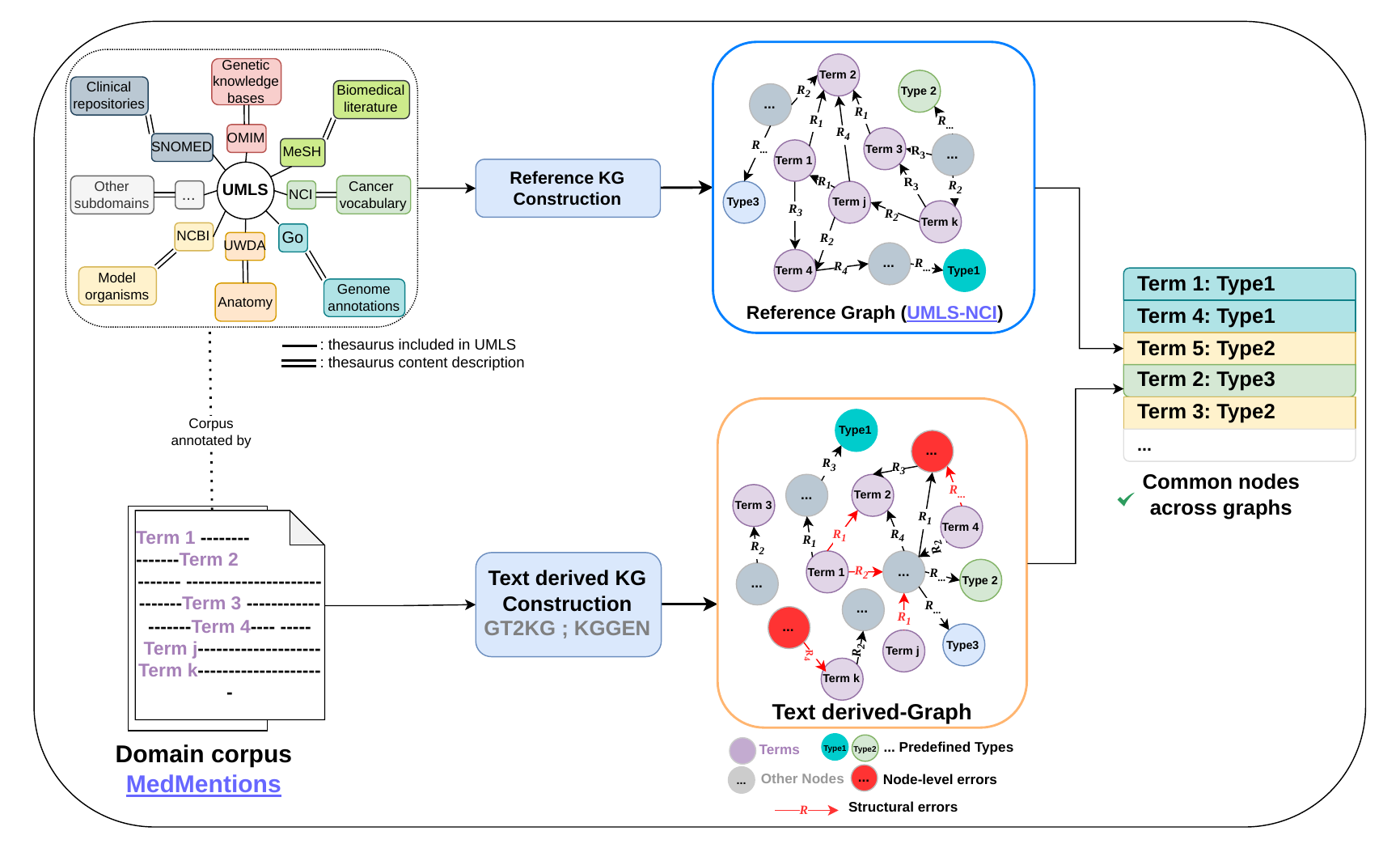}
  \caption{Overview of the Benchmark Construction Methodology.}
  \label{fig:dual_eval}
\end{figure*}

In this section, we present the construction process of our dual-purpose benchmark, which is designed to support controlled downstream evaluation by providing multiple knowledge graphs built from the same underlying corpus using different automatic construction methods.
Specifically, we generate two text-derived knowledge graphs using two different extraction pipelines, resulting in graphs that vary in structure and noise characteristics. In parallel, we construct a high-quality reference knowledge graph that shares a common set of entities with the automatically generated graphs and follows the same annotation scheme.
An overview of the benchmark construction methodology is illustrated in Figure~\ref{fig:dual_eval}.
\subsection{Reference Knowledge Graph: UMLS-NCI}
\begin{figure*}[t]
    \centering
    
    \begin{subfigure}[t]{0.58\linewidth}
        \centering
        \includegraphics[width=\linewidth]{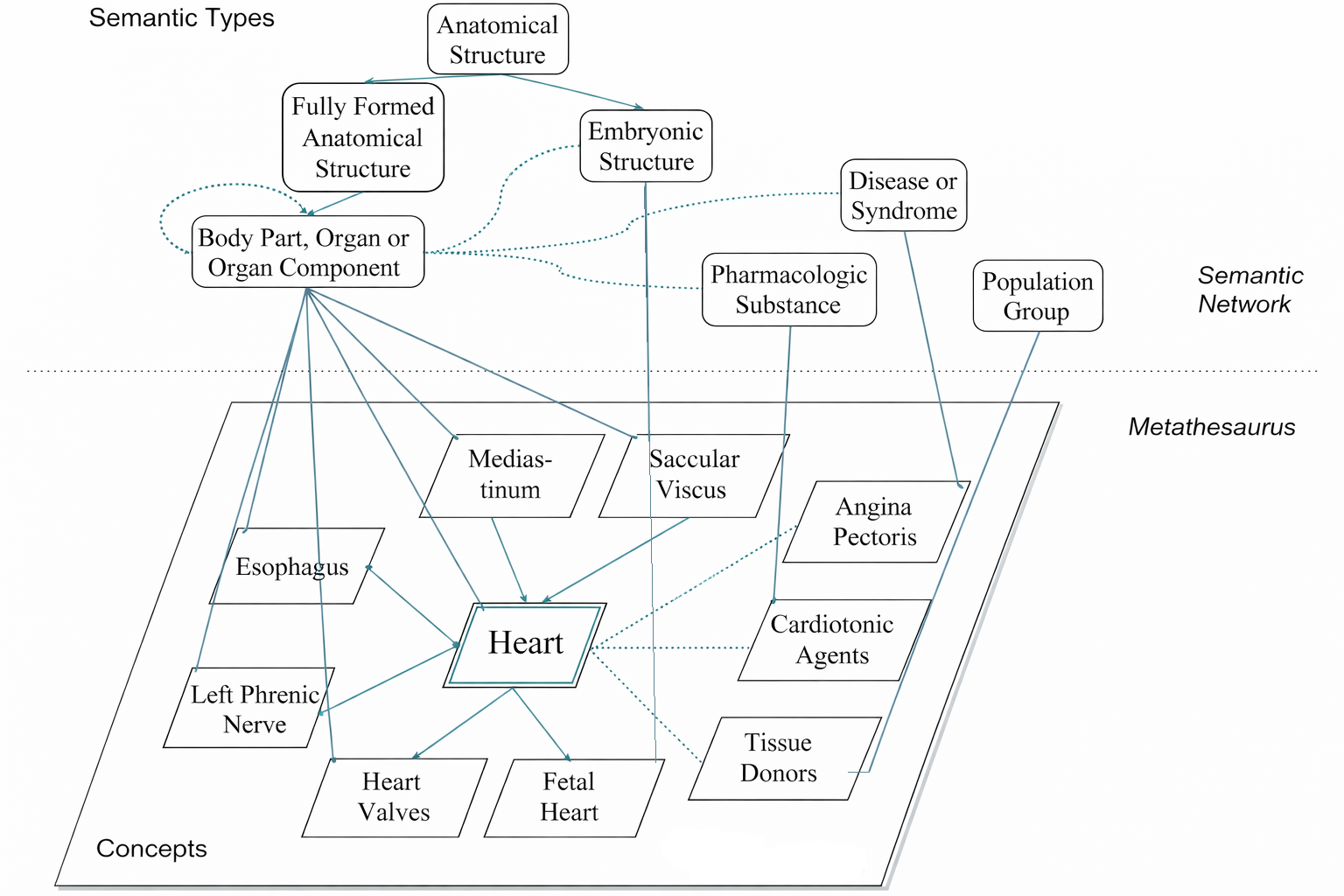}
        \caption{Relationship between the Semantic Network and the Metathesaurus.}
        \label{fig:semantic_network_example_a}
    \end{subfigure}
    \hfill
    \begin{subfigure}[t]{0.38\linewidth}
        \centering
        \includegraphics[width=\linewidth]{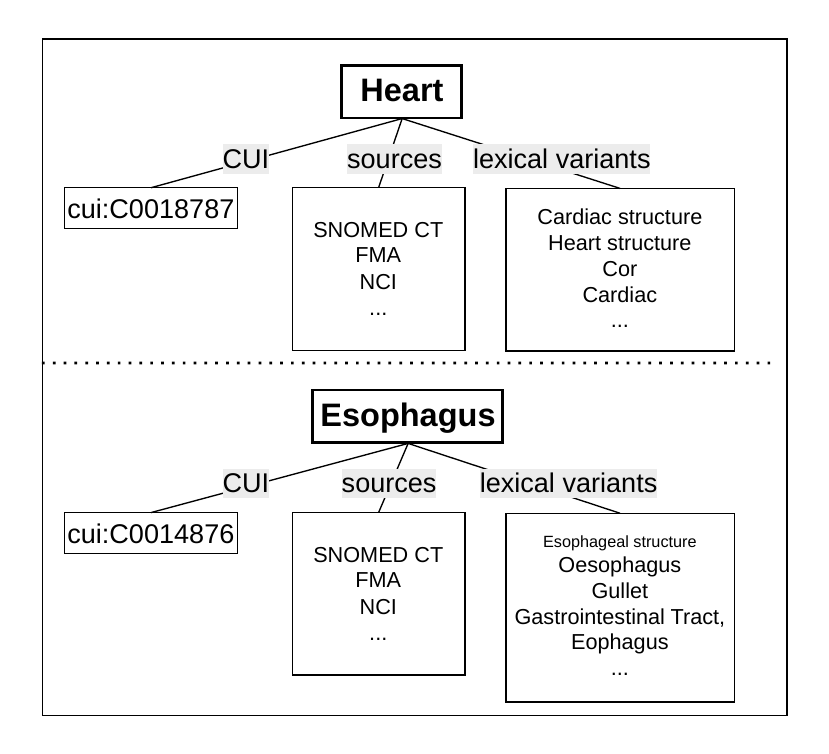}
        \caption{Examples of canonical names and lexical variations.}
        \label{fig:semantic_network_example_b}
    \end{subfigure}
    
    \caption{Illustration of the relationship between the Semantic Network and the Metathesaurus, adopted from~\cite{bodenreider2004unified}, together with examples of canonical names and lexical variations for biomedical concepts. In the Semantic Network, solid arrows represent \textit{is-a} relations, while dashed arrows denote other semantic relations between semantic types. At the Metathesaurus level, edges correspond to relations between concepts, which may include \textit{is-a} as well as other types of relations. Links between the two layers indicate the assignment of each concept to its corresponding semantic type. The right panel shows examples of canonical names and source-specific lexical variations associated with two concepts.}
    \label{fig:semantic_network_example}
\end{figure*}

\begin{figure*}[ht]
  \centering
  \includegraphics[width=\linewidth]{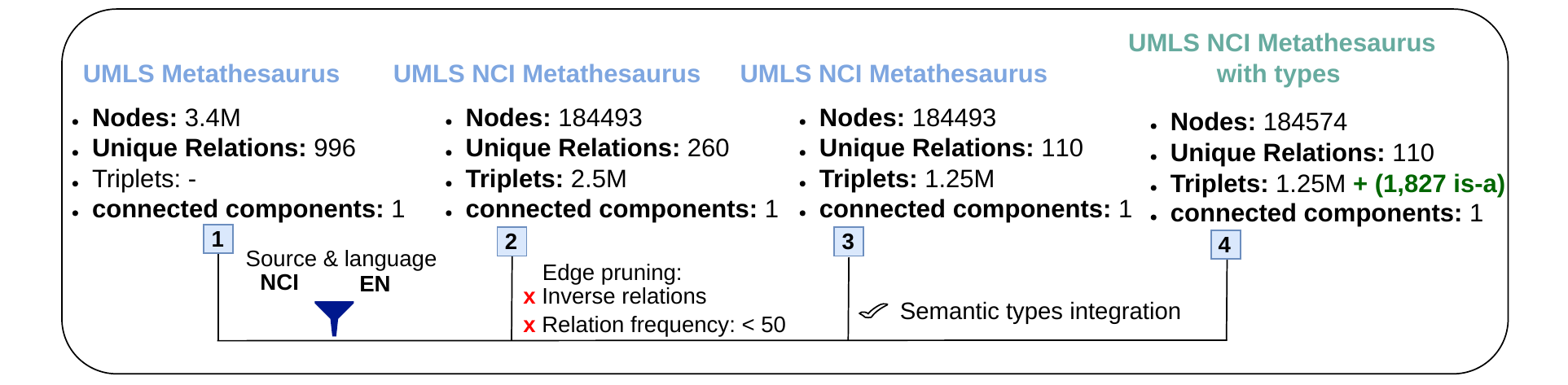}
  \caption{Stepwise construction of the reference graph (UMLS-NCI), showing the graph statistics after each filtering and preprocessing operation}
  \label{fig:umls_pipeline}
\end{figure*}

\begin{figure*}[htbp]
    \centering
    \includegraphics[width=\linewidth]{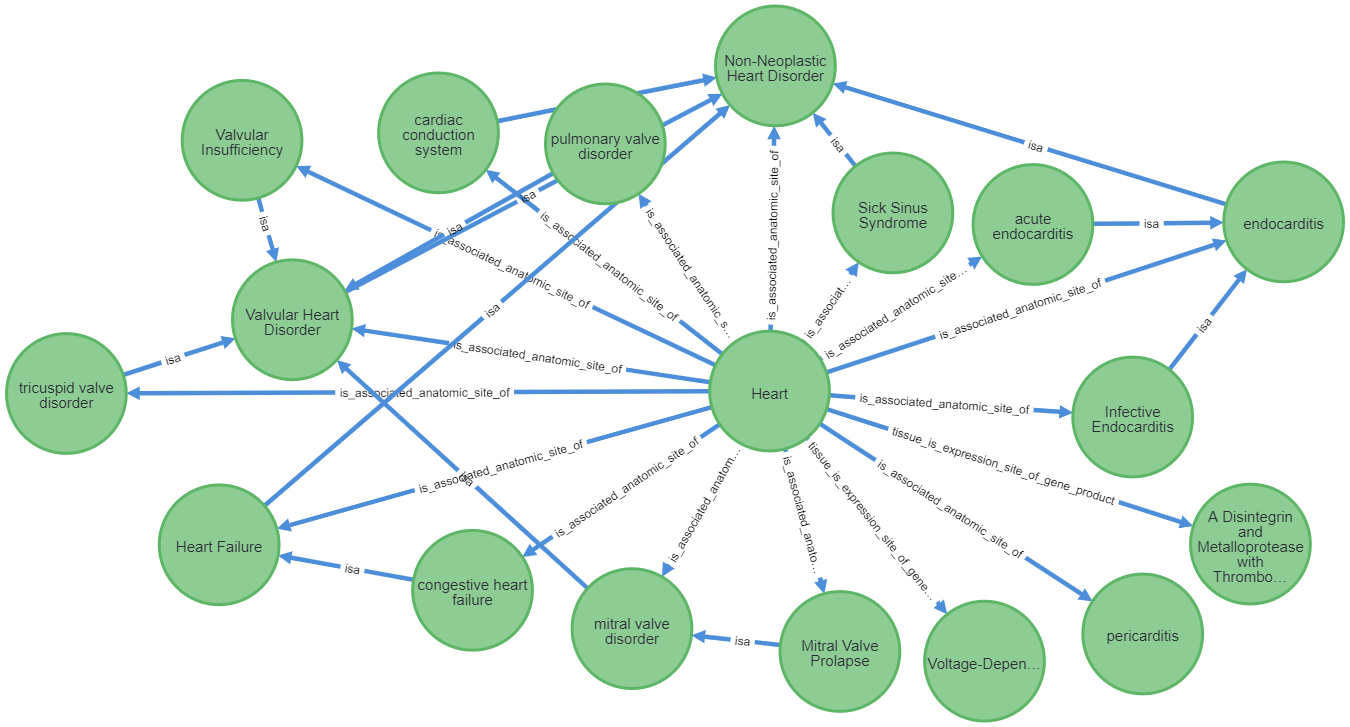}
    \caption{Example of a subgraph extracted from the reference graph, illustrating the local neighborhood of the node “Heart” and the semantic relations connecting it to related biomedical concepts.}
    \label{fig:reference-subgraph}
\end{figure*}

The reference knowledge graph (\(G_{\text{ref}}\)) is derived from the \textit{Unified Medical Language System (UMLS, 2024 release\footnote{http://www.nlm.nih.gov/research/umls/licensedcontent/umlsknowledgesources.html})}~\cite{bodenreider2004unified}, a well-established, expert-curated biomedical resource recognized for its comprehensive coverage, structural rigor, and semantic richness. UMLS is particularly suited for benchmarking due to its detailed annotation schema and abundant textual information at both the node and relation levels, facilitating a wide range of downstream applications. UMLS comprises three principal components:

\begin{itemize}
    \item \textbf{Metathesaurus}: A large repository of normalized biomedical concepts aggregated from multiple terminologies, including lexical variants and concept identifiers.
    \item \textbf{Semantic Network}: A hierarchy of 123 predefined semantic types, each linked to concepts in the Metathesaurus.
    \item \textbf{Lexical Tools}: A set of utilities for standardizing and extracting lexical forms from text.
\end{itemize}
The Metathesaurus serves as the foundation for constructing the reference graph.
Each biomedical concept is associated with a canonical name and lexical variants, 
and is typed according to the Semantic Network, as illustrated in Figure~\ref{fig:semantic_network_example}.
% \begin{figure*}[t]
%     \centering
%     \includegraphics[width=0.85\linewidth]{figures/sm_metathes_2.png}
%     \caption{Illustration of the relationship between the Semantic Network and the Metathesaurus, adopted from~\cite{bodenreider2004unified}. 
% In the Semantic Network, solid arrows represent \textit{is-a} relations, while dashed arrows denote other semantic relations between semantic types. 
% At the Metathesaurus level, edges correspond to relations between concepts, which may include \textit{is-a} as well as other types of relations. 
% Links between the two layers indicate the assignment of each concept to its corresponding semantic type.}
%     \label{fig:semantic_network_example}
% \end{figure*}
Relations between concepts are explicitly defined, each annotated with semantic predicates. This structured representation ensures that both node and edge information can be semantically interpreted and grounded in language, enabling integration with PLMs and other text-centric models.
However, the full UMLS graph is extremely large, containing millions of nodes and edges and structurally heterogeneous, exhibiting redundant or overlapping relations and inconsistent granularity in semantic types. These characteristics present challenges for graph-based learning, particularly in the context of GNN training, where scalability, homogeneity, and label consistency are essential.
To mitigate these issues, we curated a focused subgraph from the UMLS Metathesaurus using the \textit{NCI (National Cancer Institute) Thesaurus} as the primary source vocabulary, applying the adaptations illustrated in Figure~\ref{fig:umls_pipeline}.
\begin{itemize}
    \item \textit{Source and language filtering:} The UMLS Metathesaurus aggregates biomedical concepts from a diverse array of heterogeneous sources, 
    % resulting in an extremely large and terminologically inconsistent graph. 
    resulting in an extremely large graph with substantial terminological variation.
    Following prior work that treats each source vocabulary as a distinct dataset~\cite{babaei2023llms4ol}, we restricted our subgraph to a single source \textit{NCI Thesaurus} to maintain semantic coherence and facilitate manageable graph size. The NCI Thesaurus was selected for its domain relevance, clearly defined type assignments. Furthermore, we filtered the graph to retain only English-language concepts. This step reduced the graph size from over 3.4 million concepts to approximately 180,000 nodes and 1 million relations.
    
    \item \textit{Edge pruning:} Despite source filtering, the resulting graph remained densely connected and exhibited a highly skewed distribution of relation types. To mitigate this imbalance, we applied two pruning strategies. First, we removed all inverse edges, consistent with best practices in benchmarks such as FB15k-237~\cite{toutanova2015observed}. Second, we excluded relation types with fewer than 50 occurrences, as their contribution is negligible compared to high-frequency relations. 

 \item \textit{Explicit type integration:} In UMLS, semantic types are stored separately from the concept network, limiting their immediate usability for downstream graph-based learning. To enable semantic-aware modeling, we explicitly integrated the \textit{Semantic Network} into the graph by adding a dedicated node for each semantic type. Each type node was linked via \texttt{is-a} relations to a randomly selected 1\% subset of its associated concepts. 
    % This enrichment introduces hierarchical type information into the graph and supports tasks such as semantic integration and type-aware representation learning~\cite{hao2019universal}.
    This design choice is motivated by two reasons. First, in text-derived graphs, term-type relations naturally emerge, particularly because in natural language, terms are frequently introduced through definitions or contextual explanations using \textit{is-a} constructions (e.g., "X is a Y"), which explicitly associate a term with its type. Such patterns are commonly captured by information extraction pipelines, making term-type relations a ubiquitous and structurally important feature of text-derived graphs. Second, this integration injects hierarchical type information directly into the graph, supporting semantic integration and type-aware representation learning~\cite{hao2019universal}.
\end{itemize}
% The statistics of the resulting reference graphs are summarized in Table~\ref{tab:dataset-stats}. 
An example of a reference subgraph is illustrated in Figure~\ref{fig:reference-subgraph}.
\subsection{Text Corpus: MedMentions}
\begin{figure*}[t]
    \centering
    \includegraphics[width=0.92\linewidth]{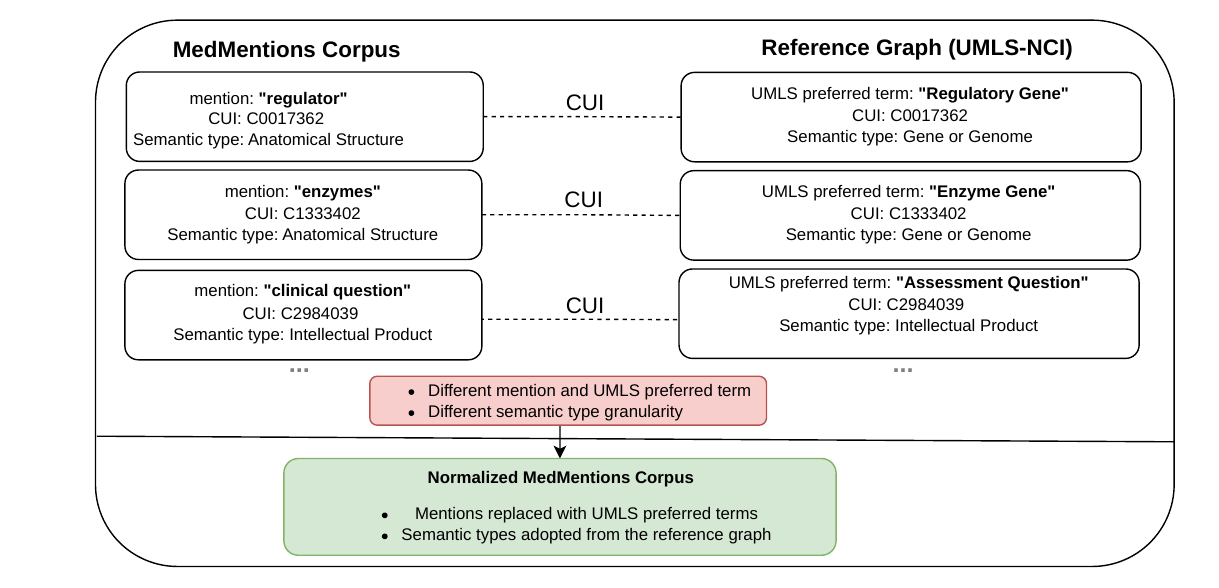}
    \caption{Examples of inconsistencies between MedMentions mentions and UMLS-NCI concepts, including differences in surface forms and semantic type granularity, motivating the normalization of the corpus before graph construction.}
    \label{fig:mm-normalization}
\end{figure*}

The construction of automatically derived knowledge graphs requires a textual corpus \(C\) containing terms that are aligned with concepts from the reference graph \(G_{ref}\). This alignment is essential, as it establishes a common set of entities between the text-derived graph and the reference graph, thereby enabling meaningful comparison and controlled evaluation across graph construction methods.
The \textbf{MedMentions} (MM) dataset~\cite{mohan2019medmentions}, designed specifically for biomedical concept recognition, is well-suited for this purpose. It consists of 4,392 PubMed~\footnote{https://pubmed.ncbi.nlm.nih.gov} abstracts, each annotated by domain experts with biomedical concept mentions mapped to UMLS concept identifiers (CUIs). We adopt MM as the textual source corpus for constructing our benchmark graphs.
However, a preprocessing step is necessary because the same UMLS-NCI concept can be associated with different textual mentions expressing the same concept through different surface forms. For example, as shown in Figure~\ref{fig:mm-normalization}, the term “regulator” in MM is aligned with concept C0017362, whose preferred term in UMLS-NCI is “gene regulatory”. In the context of MedMentions, this term clearly refers to gene regulators but is expressed differently in the text.
Additionally, at the semantic type level, MedMentions tends to use broader and more general semantic types, whereas UMLS-NCI provides more fine-grained classifications. For instance, C0017362 is typed as ``Gene or Genome'' in UMLS, whereas in MM it is typed using its more general parent type,"Anatomical Structure".

To resolve these inconsistencies, we replaced all mentions in the MM corpus with their corresponding UMLS preferred term. This normalization step ensures a consistent vocabulary, enabling reliable identification of overlapping terms and shared structure between the reference graph and text-derived graphs.

\subsection{Text-Derived Knowledge Graphs}
This step involves leveraging the prepared corpus to construct multiple knowledge graphs using distinct extraction pipelines. Each resulting graph serves as an independent dataset within the benchmark, enabling a controlled comparison of graph construction methodologies under identical textual conditions.
In this benchmark, we construct two graphs using two domain-agnostic graph construction approaches:
\subsubsection{GT2KG Graph}
Constructed using the GT2KG~\cite{kabal2024enhancing} method, this graph is generated by transforming textual sentences into subject-predicate-object triples through an OpenIE pipeline, followed by syntactic cleaning and LLM-based validation to filter out incorrect triples. 
% Summary statistics of the resulting graph are reported in Table~\ref{tab:dataset-stats}.
As expected for automatically extracted knowledge graphs, the resulting structure exhibits several imperfections.
Structurally, it shows weak connectivity, with an average directed degree of approximately 1. It is composed of multiple connected components, including one large component and several smaller ones.\\ At the edge level, the graph contains various inaccuracies such as incorrect semantic directionality, misinterpreted relation predicates, or even entirely spurious assertions. These issues distort both the relational structure and the semantics of the graph. At the node level, additional problems arise, including overly generic terms, incomplete expressions, and noisy phrases with irrelevant modifiers. Representative examples of such edge- and node-level noise are provided in Table~\ref{tab:edge_noise_examples}. 
\begin{table}[t]
\footnotesize
\centering
\setlength{\tabcolsep}{3pt}
\caption{Examples of real-world noise in the GT2KG text-derived graph.}
\label{tab:edge_noise_examples}
\begin{tabularx}{\columnwidth}{p{0.38\columnwidth} p{0.14\columnwidth} X}
\toprule
\textbf{Extracted triplet} & \textbf{Noise type} & \textbf{Explanation} \\
\midrule
(heavy metal, \textit{is}, abiotic stress) 
& Edge 
& Incorrect \textit{is-a} relation: heavy metals may cause stress but are not a subtype of it. \\

(Kit method, \textit{solve}, Cell Survival) 
& Edge 
& Incorrect predicate: the method measures survival; \enquote{solve} misrepresents the relation. \\

(mesenchymal stem cell MTH-68H, \textit{provide}, Newcastle disease virus novel therapeutic approach) 
& Node 
& The object mixes a valid concept (virus) with an irrelevant phrase, \enquote{novel therapeutic approach}. \\

(purpose of retrospective study, \textit{analyse}, incidence) 
& Node 
& The nodes are overly abstract and lack specificity. \\

(median age, \textit{is}, 63 year) 
& Node+ Edge 
& The edge should express a value relation rather than \textit{is-a}, and the numeric expression is not a proper entity. \\
\bottomrule
\end{tabularx}
\end{table}
\subsubsection{KGGen Graph}
The KGGen graph was constructed using the KGGen~\cite{mo2025kggen} method, a fully LLM-based pipeline comprising two primary stages. In the first stage, entity-relation triplets are extracted from text through prompt-based LLM querying. The second stage addresses graph sparsity and fragmentation through entity and relation resolution.
We used DeepSeek-Chat as the underlying language model and processed abstracts sequentially, one at a time. This procedure produced an initial graph containing approximately 168,797 triplets, 93,109 unique entities, and 35,178 distinct predicates. To reduce redundancy and limit LLM query costs, we applied the \texttt{DeduplicateMethod.LM\_BASED} deduplication strategy, resulting in 56,315 unique entities and 5,787 distinct predicates, which remains a relatively large number. 
% Summary statistics of the resulting graph are reported in Table~\ref{tab:dataset-stats}.
Although the KGGen graph contains substantially more entities and triplets than the GT2KG graph, it still exhibits several imperfections. Structurally, the graph remains fragmented, with multiple disconnected components and a wide diversity of relation types. This characteristic presents challenges for downstream applications such as training relational graph neural networks (e.g., RGCN~\cite{10.1007/978-3-319-93417-4_38}), which require a separate parameter matrix for each relation type. Unlike GT2KG, which maps relations to a predefined dictionary, KGGen operates in an open-schema setting. Furthermore, this graph is not immune to the typical noise encountered in LLM-generated knowledge graphs. Errors include incorrect relation types, overly generic or composite entities, and ambiguous phrase boundaries. Examples of such structural and semantic noise are illustrated in Table~\ref{tab:edge_noise_examples2}.
\begin{table}[t]
\footnotesize
\centering
\setlength\extrarowheight{2pt}
\caption{Examples of real-world noise in the KGGEN text-derived graph.}
\label{tab:edge_noise_examples2}
\begin{tabular}{p{2.7cm}p{1.5cm}p{3.2cm}}
\toprule
\textbf{Extracted triplet} & \textbf{Noise type} & \textbf{Explanation} \\
\midrule
(psa, \textit{is also known as}, psa) 
& Node 
& Entity resolution error: both nodes refer to the same concept, creating a meaningless self-relation. \\

(37 adults, \textit{did not have}, psychiatric disorders) 
& Node 
& The nodes are overly generic and population-based; ``37 adults'' is a raw count rather than a proper biomedical entity. \\

(Complication, \textit{of}, Behcet's syndrome) 
& Edge 
& ``of'' is not a valid semantic predicate; it expresses a grammatical link rather than an explicit biomedical relation. \\

(remission, \textit{or}, complete response) 
& Edge 
& ``or'' is a logical connector, not a semantic relation, and therefore cannot represent a meaningful edge. \\

(study, \textit{is a}, placebo-controlled study) 
& Node + Edge 
& The node ``study'' is too generic, and the \textit{is-a} relation is incorrectly oriented. \\
\bottomrule
\end{tabular}
\end{table}
\subsection{Common Nodes Across Graphs}
\label{sec:common_nodes}
Once the different graphs $G_{\text{ref}}$, GT2KG, and KGGEN were constructed, this step consists of identifying the nodes shared across all graphs and building an annotated dataset suitable for downstream evaluation tasks.
Since the UMLS Semantic Network defines 123 semantic types at varying levels of granularity, directly using all types would be impractical and potentially confusing. Therefore, we applied a multi-criteria filtering process, inspired by MedMentions~\cite{mohan2019medmentions}, to reduce the number of semantic types while ensuring a sufficiently general level of granularity.

\begin{figure*}[htbp]
    \centering
    \includegraphics[width=1.08\linewidth]{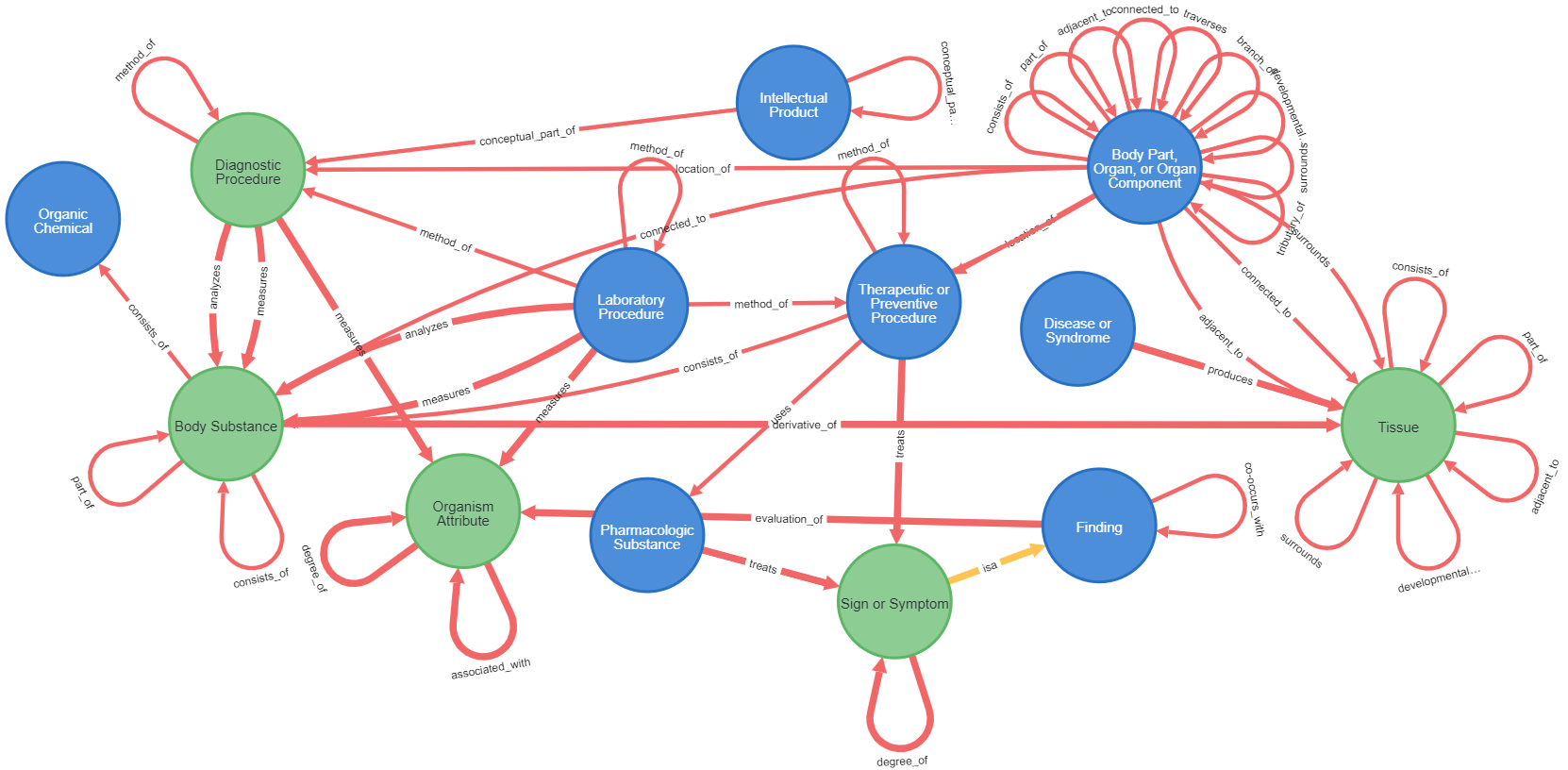}
    \caption{Illustration of a subgraph of the Semantic Network schema for the selected semantic types (shown in blue) and other semantic types (shown in green), highlighting the semantic relations between them.}
    \label{fig:semantic-network}
\end{figure*}

The filtering procedure was defined as follows:
\begin{itemize}
    \item \textbf{Type frequency:} We retained only semantic types with at least 2{,}000 instances in the reference graph.
    \item \textbf{Intersection and final filtering:} To ensure compatibility across the three graph variants, we intersected their node sets. We first intersected the cleaned reference graph with the GT2KG graph, which contains fewer entities than KGGen. This intersection was performed using exact string matching and yielded approximately 2{,}000 shared nodes.  
    To ensure a balanced representation across semantic types, we applied an additional frequency constraint, discarding semantic types with fewer than 100 shared nodes. This filtering step resulted in 1{,}040 nodes spanning eight semantic types common to both the reference and GT2KG graphs.  
    Subsequently, we aligned these nodes with their counterparts in the KGGen graph. Due to variability in entity surface forms, we employed exact string matching and adapted KGGen’s entity resolution mechanism. For each cluster of coreferent entities, we selected as the canonical representative the entity form appearing in the 1{,}040 node set whenever possible.
\end{itemize}
Following this procedure, we obtained a final set of \textbf{eight semantic types}, comprising a total of \textbf{1,032} aligned nodes across the three graphs. The organization of these semantic types within the semantic network is illustrated in Figure~\ref{fig:semantic-network}.
Table~\ref{tab:dataset-stats} summarizes all structural statistics of the graphs provided in our benchmark.
\begin{table}[t]
\small
\centering
\caption{Summary of the structural statistics of all graphs provided in the benchmark, including the reference graph \(G_{\text{ref}}\) (UMLS-NCI) and the automatically constructed GT2KG and KGGEN graphs.}
\label{tab:dataset-stats}
\begin{tabular}{p{3.1cm}ccc}
\toprule
\textbf{Metric} 
& \shortstack{\textbf{\(G_{\text{ref}}\)}\\(UMLS-NCI)} 
& \shortstack{\textbf{GT2KG}\\\textbf{Graph}} 
& \shortstack{\textbf{KGGEN}\\\textbf{Graph}} \\
\midrule
Number of nodes        & 184,574   & 37,156  & 56,315  \\
Number of edges        & 1,258,931 & 34,322  & 104,223 \\
Unique relations       & 110       & 76      & 5,787   \\
Connected components   & 1         & 6,987   & 2,551   \\
Largest component size & 184,574   & 19,028  & 50,324  \\
\midrule
\multicolumn{4}{l}{\textbf{Common nodes across graphs}} \\
Annotated nodes        & \multicolumn{3}{c}{1,032} \\
Number of classes      & \multicolumn{3}{c}{8} \\
\bottomrule
\end{tabular}
\end{table}
\section{Evaluation Protocol}
\label{sec:evaluation_protocol}
\begin{figure*}[ht]
  \centering
  \includegraphics[width=\linewidth]{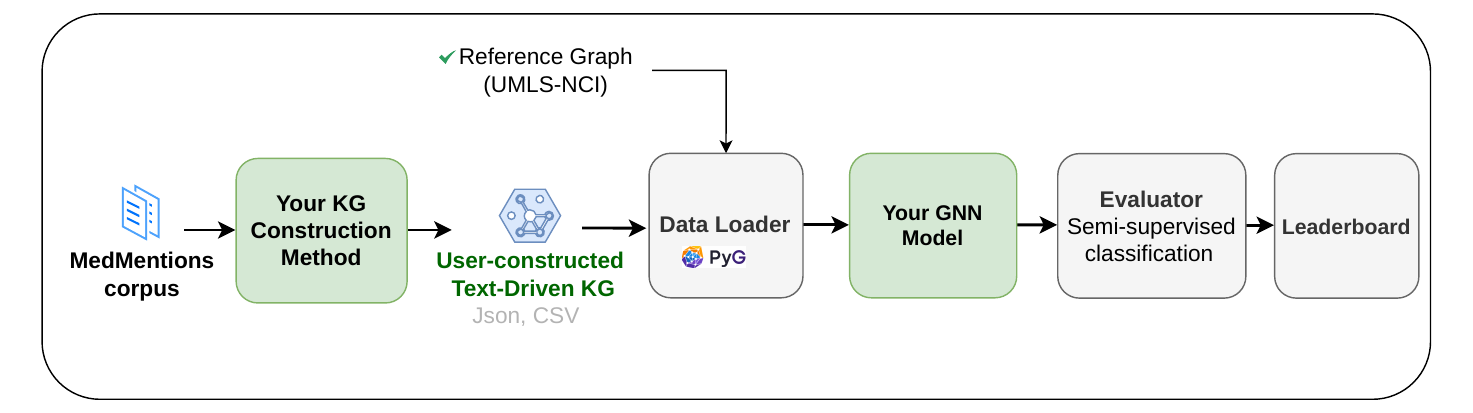}
  \caption{Overview of the evaluation pipeline. Green components correspond to user-defined inputs.}
  \label{fig:evaluation_pipeline}
\end{figure*}

To achieve the dual objective of this benchmark, namely evaluating (i) automatic knowledge graph construction methods on a downstream task under controlled experimental conditions, and (ii) GNN models operating on text-derived knowledge graphs. We provide a fully automated evaluation framework that standardizes the training and evaluation process across all models and graphs.
The benchmark supports two distinct evaluation scenarios, as illustrated in Figure~\ref{fig:evaluation_pipeline}:\\
\textbf{Scenario 1: Evaluating Knowledge Graph Construction Methods.} 
Researchers developing new knowledge graph extraction pipelines can evaluate their methods by applying them to the MedMentions corpus and generating a corresponding knowledge graph. 
The constructed graph is expected to cover the set of the common terms. If some common nodes are missing, the evaluation is restricted to the subset of nodes present in the graph. However, incomplete coverage directly reflects a limitation of the construction method, as it indicates missed relevant entities (i.e., lower recall).
The constructed graph is then evaluated through a downstream semi-supervised classification task, using GNN models under a standardized evaluation protocol. This enables direct comparison with graphs produced by existing construction methods, as well as with the clean reference graph, which serves as an upper bound.
Finally, graphs generated by new methods can be integrated into the benchmark as additional datasets,  making the benchmark continuously extensible.\\
\textbf{Scenario 2: Evaluating GNN Models.} 
Researchers developing new GNN architectures can evaluate their models on any of the provided graphs (Reference, GT2KG, or KGGen). To facilitate integration, users are only required to implement the forward pass of their model following a simple and standardized interface.
The benchmark framework handles the entire training and evaluation pipeline, including data loading, hyperparameter configuration, multi-seed training, and metric computation.

\smallskip
\noindent\textit{Note: For consistency, the triplets introduced during the semantic type integration step in the clean reference graph can also be added to the automatically constructed graphs. This ensures a controlled comparison setting. However, this augmentation is optional and left to the user’s discretion, allowing evaluation both with and without type-level structural enrichment.}
\subsection{Data Loader}
All graphs included in the benchmark are encapsulated within a \texttt{PyTorch Geometric Dataset} object to ensure compatibility and ease of use. The data loader is configurable with respect to the node and relation vectorization strategy, allowing users to choose between random initialization or any PLM available through the HuggingFace ecosystem. This design enables systematic analysis of the impact of PLM-based initialization on downstream performance. 
The data loader also supports newly generated graphs produced by additional extraction methods. Integrating a new graph only requires adhering to the prescribed file structure, as detailed in the repository documentation~\footnote{~\url{https://github.com/OthmaneKabal/text_driven_kg_bench}}. An example usage is shown below:
\begin{center}
\begin{verbatim}
tdg = TDGBench()
t2kg_graph_dataset = tdg.get_data(
    kg_name="GT2KG_Graph",
    init_embd="all-MiniLM-L6-v2",
    semantic_type_integrated=True
)
\end{verbatim}
\end{center}
\subsection{Evaluator}
We provide a unified evaluator responsible for standardized training and evaluation across all benchmark graphs, as well as compatibility with newly generated graphs that follow the prescribed data format.
Below are two example usage scenarios:
\begin{center}
\begin{minipage}[t]{0.45\textwidth}
\textbf{Evaluating a new generated graph with GNN baselines:}
\begin{verbatim}
tdg = TDGBench()
tdg.evaluate(
    kg_name="new_kg",
    GNN_model="Baselines",
    init_embd="all-MiniLM-L6-v2",
    semantic_type_integrated=True
)
\end{verbatim}
\end{minipage}
\hfill
\begin{minipage}[t]{0.45\textwidth}
\textbf{Evaluating the reference graph with a custom GNN:}
\begin{verbatim}
tdg = TDGBench()
tdg.evaluate(
    kg_name="reference graph",
    GNN_model=new_GNN_model,
    init_embd="all-MiniLM-L6-v2",
    semantic_type_integrated=True
)
\end{verbatim}
\end{minipage}
\end{center}
\subsubsection{Task Definition and Data Splits}
Evaluation is conducted on the \textbf{semi-supervised node classification task}, where GNN models are trained using a limited subset of labeled nodes and evaluated on a held-out test set. This task is particularly relevant for text-derived knowledge graphs, where annotation is expensive and models must leverage both labeled nodes and graph structure for effective learning.\\
Performance is measured using standard classification metrics computed on the annotated nodes: \textbf{Accuracy, Macro F1-score, Macro Precision and Recall.}\\
For each graph, the 1,032 labeled nodes are split into training, validation, and test sets using a \textbf{10/10/80 ratio}. %(approximately 103 training, 103 validation, and 826 test nodes).
The splits are \textbf{stratified by type} to preserve class proportions across sets and avoid class imbalance. To ensure reproducibility, we provide \textbf{pre-generated splits} with fixed random seeds [\texttt{seeds: 42, 123, 456, 789, 1000}]. Unless otherwise specified, we report the \textbf{mean and standard deviation} over these $N=5$ random splits. All split files are released alongside the benchmark.
\subsubsection{Standardized Training Protocol}
We provide a standardized training framework for all GNN models in order to eliminate variations arising from implementation-specific choices. The framework also allows users either to employ their own classifier or to use the one we provide, namely an output MLP layer with 8 units corresponding to the target classes. Training is performed using the Adam optimizer with a learning rate of 0.01 and an L2 regularization term with a weight decay of $5 \times 10^{-4}$. The maximum number of training epochs is set to 200, with early stopping based on validation loss and a patience of 20 epochs. A dropout rate of 0.5 is applied between layers to mitigate overfitting. Hidden dimensions are selected from $\{64, 128, 256\}$ through hyperparameter optimization. Mini-batch training is conducted using neighborhood sampling following the inductive learning approach of~\cite{hamilton2017inductive}, with 200 neighbors sampled per hop. By default, node and relation embeddings are initialized using the pretrained PLM \texttt{all-MiniLM-L6-v2}.\\
Model selection is performed based on validation data. Each training procedure is repeated five times using different random seeds (\texttt{42, 123, 456, 789, 1000}), and the reported results correspond to the mean and standard deviation across runs in order to ensure reproducibility.

\subsection{Leaderboard}
To promote transparent and reproducible evaluation, we maintain a public leaderboard hosted on Hugging Face Spaces~\footnote{https://huggingface.co/spaces/othmanekabal/tdg-bench}. This leaderboard displays performance metrics for all baseline models included in the benchmark, as well as community-submitted models.
\section{Baselines}
To provide strong and reproducible reference points for evaluation, we implement several widely adopted GNN architectures as baseline models. These baselines serve both as diagnostic tools to assess the intrinsic difficulty of the proposed benchmark and as standardized reference methods for comparing new knowledge graph construction pipelines or novel GNN architectures.
We select representative models covering complementary message-passing paradigms, including convolution-based, attention-based, and relation-aware approaches:
\begin{itemize}
    \item \textbf{GCN}~\cite{kipf2016semi}, which performs spectral graph convolution with shared weights across neighbors;
    \item \textbf{GAT}~\cite{velivckovic2017graph}, which introduces learnable attention coefficients to adaptively weight neighbor contributions;
    \item \textbf{RGCN}~\cite{10.1007/978-3-319-93417-4_38}, specifically designed for multi-relational graphs through relation-specific transformations;
    \item \textbf{TransGCN / RotatE-GCN}~\cite{cai2019transgcn}, which integrate geometric relation transformations with graph convolution to better capture structured relational patterns. For these models, we evaluate both standard and attention-enhanced variants.
\end{itemize}
For fairness, all baselines share the same architectural design and training protocol. Each model consists of two message-passing layers, with dropout applied between layers to mitigate overfitting. The final classification layer is provided by the benchmark evaluator to ensure consistency across models. Training strictly follows the standardized framework described in the previous \autoref{sec:evaluation_protocol}, including identical data splits, optimization settings, hyperparameter search space, and PLM-based node initialization. The baseline results are summarized in Table~\ref{tab:baseline_results}.
\newcommand{\mstd}[2]{\ensuremath{#1_{\scriptsize\pm\,#2}}}
\newcommand{\mstdB}[2]{\ensuremath{\mathbf{#1}_{\scriptsize\pm\,#2}}}
\newcommand{\attn}{\ensuremath{_{\scriptsize\text{attn}}}}
\newcommand{\conv}{\ensuremath{_{\scriptsize\text{conv}}}}
\begin{table*}[!t]
\small
\centering
\setlength{\tabcolsep}{5pt}
\caption{Results of the baselines. The best results in each graph are shown in bold. `--' indicates that the model was not evaluated.}
\label{tab:baseline_results}
\begin{tabular}{l cc cc cc}
\toprule
& \multicolumn{2}{c}{\textbf{Reference ($G_{\text{ref}}$})} 
& \multicolumn{2}{c}{\textbf{GT2KG}} 
& \multicolumn{2}{c}{\textbf{KGGEN}} \\
\cmidrule(lr){2-3}\cmidrule(lr){4-5}\cmidrule(lr){6-7}
\textbf{Model} & \textbf{Acc.} & \textbf{F1} & \textbf{Acc.} & \textbf{F1} & \textbf{Acc.} & \textbf{F1} \\
\midrule
GCN 
  & \mstd{0.662}{0.011} & \mstd{0.655}{0.008}
  & \mstd{0.392}{0.016} & \mstd{0.390}{0.026}
  & \mstd{0.367}{0.019} & \mstd{0.364}{0.021} \\
GAT 
  & \mstd{0.682}{0.012} & \mstd{0.675}{0.012}
  & \mstd{0.425}{0.012} & \mstd{0.422}{0.023}
  & \mstd{0.385}{0.017} & \mstd{0.380}{0.015} \\
RGCN 
  & \mstdB{0.712}{0.009} & \mstdB{0.708}{0.007}
  & \mstd{0.508}{0.020} & \mstd{0.499}{0.019}
  & -- & -- \\
TransGCN\conv 
  & \mstd{0.696}{0.010} & \mstd{0.691}{0.007}
  & \mstd{0.508}{0.017} & \mstd{0.502}{0.019}
  & \mstd{0.571}{0.024} & \mstd{0.567}{0.030} \\
TransGCN\attn 
  & \mstd{0.645}{0.022} & \mstd{0.642}{0.023}
  & \mstd{0.504}{0.019} & \mstd{0.501}{0.023}
  & \mstd{0.578}{0.020} & \mstd{0.573}{0.015} \\
RotatE-GCN\conv 
  & \mstd{0.673}{0.020} & \mstd{0.669}{0.019}
  & \mstd{0.542}{0.018} & \mstd{0.534}{0.022}
  & \mstd{0.576}{0.023} & \mstd{0.569}{0.021} \\
RotatE-GCN\attn 
  & \mstd{0.668}{0.024} & \mstd{0.664}{0.022}
  & \mstdB{0.565}{0.021} & \mstdB{0.563}{0.034}
  & \mstdB{0.594}{0.028} & \mstdB{0.580}{0.033} \\
\bottomrule
\end{tabular}
\end{table*}

The results highlight the combined importance of relational modeling and graph structural quality. On the reference UMLS-NCI graph, which is clean, well-connected, and semantically consistent, explicitly relational models achieve the best performance. In particular, RGCN outperforms all other approaches (Acc = 0.712, F1 = 0.708), confirming the benefit of learning relation-specific transformations in richly multi-relational settings. In contrast, simpler architectures such as GCN and GAT, which aggregate neighbors without distinguishing relation types, exhibit weaker performance.\\
When moving to the automatically extracted GT2KG graph, performance decreases for all models due to structural fragmentation and semantic noise. However, relation-aware variants such as RotatE-GCNattn and TransGCN remain noticeably more robust, with RotatE-GCNattn achieving the best score (Acc = 0.565). This trend becomes even more pronounced on KGGEN, where, despite higher noise levels, TransGCN and RotatE-GCN variants continue to outperform other approaches.\\
Their robustness can be explained by their geometric relation modeling combined with a bidirectional message passing mechanism, which propagates information in both directions along edges. This two-way aggregation improves information flow and is particularly beneficial in sparse or weakly connected graphs, where local neighborhoods provide limited signals.\\
In contrast, RGCN relies on relation-specific transformations, requiring one parameter matrix per relation. While effective on graphs with a limited number of relations, its performance degrades as the number of relation types increases, due to over-parameterization and insufficient data per relation. This limitation becomes critical in open-schema graphs such as KGGEN, where the large number of relations leads to prohibitive memory and computational costs, making RGCN impractical.\\
Overall, these results indicate that while relational models consistently outperform standard GNNs, geometric approaches such as TransGCN and RotatE-GCN are particularly well-suited for noisy, heterogeneous, and sparse graphs. Furthermore, performance depends not only on semantic quality but also strongly on structural properties such as connectivity and component size, which directly affect the effectiveness of message propagation.
\section{Conclusion}
In this paper, we introduced a novel benchmark for text-derived knowledge graphs in the biomedical domain, designed with a dual-purpose evaluation philosophy. The proposed framework simultaneously addresses two complementary objectives: (i) systematically evaluating automatic knowledge graph construction methods through their impact on a realistic downstream task, and (ii) assessing the robustness and effectiveness of Graph Neural Networks when operating on noisy, fragmented, and semantically complex graphs extracted from text.
Our main contribution lies in providing a controlled and standardized experimental environment that includes multiple graphs generated from the same textual corpus using different extraction pipelines, alongside a high-quality clean reference graph that serves as an upper performance bound. This design enables precise quantification of the performance degradation induced by structural and semantic noise, facilitates fair comparison between construction methods, and allows deeper analysis of how GNNs generalize under realistic, imperfect graph conditions.
Furthermore, the benchmark is inherently extensible. New extraction methods can be seamlessly integrated by contributing additional graphs, enabling continuous and reproducible evaluation across evolving construction approaches. This extensibility supports advanced studies on the robustness of GNN architectures and provides opportunities to investigate how structural properties (e.g., connectivity, fragmentation, and relation density) and semantic characteristics (e.g., lexical ambiguity and extraction noise) influence downstream learning performance.

Despite these strengths, several limitations should be acknowledged. First, the benchmark is currently restricted to the biomedical domain, mainly due to the difficulty and high cost of obtaining reliable clean reference graphs and consistently annotated corpora in other domains. Second, the evaluation protocol focuses on a single downstream task, namely semi-supervised node classification. Incorporating additional tasks such as link prediction or knowledge graph completion remains challenging in open-schema settings, where relation types vary substantially across automatically constructed graphs.
These limitations open promising directions for future work. Extending the benchmark to other domains would improve generalizability and broaden its applicability. Developing universal relation mapping or normalization strategies could enable additional tasks such as link prediction across heterogeneous graphs. Finally, conducting more detailed analytical studies on the relative impact of structural versus semantic graph quality could provide deeper insights into which factors most critically affect GNN performance and robustness.
\section*{Data Availability and Resources}
The code and resources associated with this benchmark are publicly available at:
\url{https://github.com/OthmaneKabal/text_driven_kg_bench}.\\
The repository includes:
\begin{itemize}
    \item Automatically constructed knowledge graphs (GT2KG and KGGen);
    \item A reference graph derived from UMLS-NCI. Due to licensing restrictions, the full UMLS Metathesaurus cannot be directly redistributed. However, we provide detailed instructions to reconstruct the reference graph from the licensed UMLS resources;
    \item The annotated node classification dataset along with predefined training, validation, and test splits;
    \item The complete evaluation pipeline, including configuration files and training scripts.
\end{itemize}

All experiments are fully reproducible using the provided scripts and fixed random seeds.

\bibliographystyle{SageH}
\bibliography{ref}

\end{document}